\documentclass[conference]{IEEEtran}
\IEEEoverridecommandlockouts
\usepackage{cite}
\usepackage{amsmath,amssymb,amsfonts}
\usepackage{algorithmic}
\usepackage{graphicx}
\usepackage{textcomp}
\usepackage{xcolor}
\usepackage{nicefrac}       
\usepackage{microtype}      
\usepackage{lipsum}
\usepackage{graphbox, graphicx}
\usepackage{subcaption}
\usepackage{multirow}
\usepackage{tabularx}
\usepackage{adjustbox}
\usepackage{makecell}
\usepackage{url}

\def\BibTeX{{\rm B\kern-.05em{\sc i\kern-.025em b}\kern-.08em
    T\kern-.1667em\lower.7ex\hbox{E}\kern-.125emX}}
\begin{document}

\title{Leveraging an Efficient and Semantic Location Embedding to Seek New Ports of Bike Share Services}

\author{\IEEEauthorblockN{Yuan Wang$^1$, Chenwei Wang$^2$, Yinan Ling$^3$, Keita Yokoyama$^2$, Hsin-Tai Wu$^2$, Yi Fang$^1$}
\thanks{*Chenwei Wang is the corresponding author.}
\IEEEauthorblockA{$^1$Dept. of Computer Science, Santa Clara University, Santa Clara, CA 95050}
\IEEEauthorblockA{$^2$Data Analytics and Innovation Group, DOCOMO Innovations Inc., Palo Alto, CA 94304}
\IEEEauthorblockA{$^3$Institute of Data Science, Columbia University, New York, NY 10027}
Email: $^1$\{ywang4, yfang\}@scu.edu, $^2$\{cwang, yokoyama, hwu\}@docomoinnovations.com, $^3$yl4030@columbia.edu}

\maketitle

\begin{abstract}
For short distance traveling in crowded urban areas, bike share services is becoming popular owing to the flexibility and convenience. To expand the service coverage, one of the key tasks is to seek new service ports, which requires to well understand the underlying features of the existing service ports. In this paper, we propose a new model, named for Efficient and Semantic Location Embedding (ESLE)\footnote{The source code of this model can be found at github.com/ywang4/ELSE-Efficient-and-Semantic-Location-Embedding.}, which carries both geospatial and semantic information of the geo-locations. To generate ESLE, we first train a multi-label model with a deep Convolutional Neural Network (CNN) by feeding the static map-tile images and then extract location embedding vectors from the model. Compared to most recent relevant literature, ESLE is not only much cheaper in computation, but also easier to interpret via a systematic semantic analysis. Finally, we apply ESLE to seek new service ports for NTT DOCOMO's bike share services operated in Japan. The initial results demonstrate the effectiveness of ESLE, and provide a few insights that might be difficult to discover by using the conventional approaches.
\end{abstract}


\section{Introduction}\label{sec:intro}

Recognized as one of the most powerful engines of the technology revolution for the next decade, deep learning has stimulated a variety of new technologies and enriched business opportunities. Using deep learning, recently  user/device location-based services and applications have attracted much attention, such as user location recommendation, trajectory prediction, and transportation. Particularly, the past a few years have witnessed the expanding of bike share services in the dense population area of many cities all over the world, because it provides a solution to the ``last mile" problem on top of the conventional transportation infrastructure. To expand the bike share service system, how to seek new service ports in a wider area has readily become a key business task. 

The location of a service port, indicated by a pair of latitude and longitude, is merely a single point on earth. While usually little interest arises for a point with Lebesgue measure zero in area, what makes the location more informative is its nearby environment that involves natural geographical information and human activities. These information can be somewhat carried by a geographical image within a certain range from the location of interest. While such an image is represented by pixels from thousands to millions scale (e.g., a $(224\times 224)$ image has 50,176 pixels per channel), processing such an image could be very computational heavy especially in many time-sensitive and storage-sensitive applications. To efficiently utilize the environment information, representation learning for location embedding (with dimension reduction) has been investigated to resolve the high-dimension challenge. In addition, due to the countless number of location points on earth, a model with sufficient scalability is highly desired. In this paper, we consider the bike share services provided by NTT DOCOMO, Inc. to the public in Japan. We are interested in the question: {\em How to design an efficient and semantic location embedding model and leverage it to seek new ports of NTT DOCOMO's bike share services?}

\subsection{Related Work}

The transformation from high-dimensional data to low-dimensional location representation is referred to as {\em location embedding}. On this avenue, there are a number of interesting models and results in the recent literature. In \cite{yin2019gps2vec}, GPS2Vec was proposed to transform GPS vector representations to 2000-dimensional vocabulary-based semantic feature vectors. They employed a multi-layer perception neural network (MLPNN) to supervise-learning the feature vectors, which were extracted from Flickr, Twitter, Foursquare. However, the semantic contexts of the collected data could be highly noisy and inaccurate. In \cite{abonce2019geolocation}, a Siamese-like embedding model was trained by using map-tile images and Google Street View images. Also, \cite{Law_2019} used convolutional autoencoder and Principal Component Analysis (PCA) \cite{jolliffe2016principal} to generate location embedding from street view images, which do not contain semantic information. Alternatively, a semi-supervised learning approach was developed to train location embedding \cite{Sentiance, urban2vec, tile2vec}. Moreover, they all chose the triplet loss as the loss function, which was successfully employed in face recognition \cite{face_recognition} to penalize the map-tile images if they are not \emph{similar} to the anchor in each triplet. Per the definition of ``similar", \cite{tile2vec, Sentiance} both assumed that physically-closed (far away) neighbors have similar (dissimilar) semantics and representations. However, for map-tile images, this assumption might be flawed. For example, in the urban dense area, the geospatial information could significantly vary, even when two locations are near to one another. For another example, the similarity was defined by POI similarity in \cite{urban2vec}. Particularly, they proposed a two-stage framework incorporating both street view imagery and point-of-interest (POI) data to learn neighborhood embedding, and thus the similarity definition is more reasonable. However, in all of \cite{Sentiance, urban2vec, tile2vec}, the triplet-loss-based models have relatively high complexity in both space and time, which are at least $O(N^2)$, where $N$ is the number of map-tile images in the dataset. Considering the scalability of the model to a very large area, i.e., $N$ is very large, all their models are weak in efficiency. 

Regarding the prediction of new service ports for bike share business, many works also provided a variety of interesting and insightful approaches, such as \cite{cetinkaya2017bike, chen2015bike, fuller2011use, rixey2013station}. However, to the best of our knowledge, location embedding has not been applied.

\subsection{The Contribution}

To overcome the complexity challenge (of the triplet loss) for scalability, we need to develop a more efficient approach for location embedding. Also, to avoid the assumption of the map-tile image similarity made in \cite{tile2vec, Sentiance}, we need a new approach to re-define the similarity. Although the POI information used in \cite{urban2vec} seems promising, it does not fully rely on the map-tile images. Instead, we need to understand the underlying features of the map-tile images. 

Per the second issue, let us re-think of what can be used for describing a location. Clearly, three types of messages -- map-tile image, meta information, POI information -- could be explored. Specifically, the map-tile image itself includes geospatial information, such as shape, color, size, etc. of the objects in the image; the meta information describes the objects in the image with numerical numbers, such as the number of roads, buildings, parks. This is typically used when one visualizes the similarity between images; and the POI information labels the functions of the objects with words or phrases. Albeit all the three types of messages are differently structured, they all describe something in the same area and do not exclude one from another. Recognizing this connection, in this paper we propose to use the meta information to label the map-tile images, which was not used in \cite{abonce2019geolocation, tile2vec, Sentiance, urban2vec, yin2019gps2vec}.

Next, back to the first issue, intuitively, a well-designed location embedding should: (1) contain a certain amount of geospatial information of the image; (2) preserve a certain amount of semantic information; (3) encode similar (dissimilar) geographical images to similar (dissimilar) embedding vectors, where the similarity between two vectors can be characterized by some metrics, such as the Euclidean distance or the cosine similarity score; (4) carry more amount of information as the number of dimensions increases. In this work, we propose a new approach to create location embedding, named for \textbf{Efficient and Semantic Location Embedding (ESLE)}. Specifically, we train a CNN \cite{lecun1995convolutional} with map-tile images as the input. Unlike \cite{Sentiance, tile2vec}, we do not consider the actual distance between the locations. Instead, we formulate a supervised multi-label classification problem by using the meta information. Such an approach not only enables to more efficiently extract the underlying features of the map-tile images but also reduces the complexity of both space and time to $O(N)$. This is because once the hyper-parameters of the model are fixed the running time and memory in needs are both linear in $N$, unlike the interplay among the components in the triplet. Therefore, compared to the semi-supervised learning approaches described in \cite{abonce2019geolocation, tile2vec, Sentiance, urban2vec}, our model is more efficient and interpretable.

Finally, we apply the obtained ESLE to analyze the data collected from NTT DOCOMO's bike share services. In particular, we first transform the locations of the existing service ports to numerical vectors. Then we analyze their underlying features and use them to seek new service ports. Based on our initial results of a variety of experiments and statistical analysis, our ESLE is able to better extract semantic/POI information, suggesting ESLE's potential as a universal tool for general location-based tasks.

\section{Data Description}\label{subsec:data_describe}

Three categories of datasets are used in this work: map-tile images, meta information, and POI information. In this section, we introduce each dataset in the following. 

\subsection{Map Tile Images}

Using the OpenStreetMap map server under zoom level 16 \cite{OpenStreetMap}, we collect a total of 1,666,168 locations in Japan in the form of (\emph{lat}, \emph{lon}). For each location, a map-tile image centered at that location is collected, and its actual coverage is automatically adjusted to 500 meters along latitude and longitude, so that these 1,666,168 map-tile images together cover the entire land territory of Japan. After a closer look, we find that a total of 743,458 of those map-tile images do not have pixel value variation, indicating either sea only or green land only. Without information variation, they do not provide much information for training a model, and thus we sample only 1,000 images for sea and another 1,000 for green lands.
Finally, the resulting number of map-tile images of interest reduces to 924,710, which is referred to as $\mathcal{M}$ for brevity throughout the paper. 

Besides the map-tile images set $\mathcal{M}$ over entire Japan, we also consider a subset of $\mathcal{M}$ to perform a number of experiments of interest. In particular, we consider the area with longitude range (139.003125, 139.996875) and latitude range (35.335417, 35.997917) that covers a actual range of $89.44 \text{km} \times 72.88 \text{km}$ on earth. Within this particular regime, there are a total of 25,600 map-tile images out of $\mathcal{M}$. For brevity, we denote this subset as $\mathcal{M}_1$.

\subsection{Metadata and Meta Label}\label{subsec:meta}

Using the OpenStreetMap Overpass API, we collect metadata describing semantic information of each map-tile image. Specifically, we consider 14 classes of meta information identified in the tuple (buildings, highway, peak, water, river, railway, rail station, park, playground, road, airport, trail, farmland, grassland). After parsing the data returned from the API, we retrieve the count number of each class for each image. The resulting 14-dimensional count vector for each image is called "metadata".

\begin{figure}[ht]
    \centering
    \begin{subfigure}[t]{.48\linewidth}
        \centering
        \includegraphics[align=t, width=1.0\linewidth]{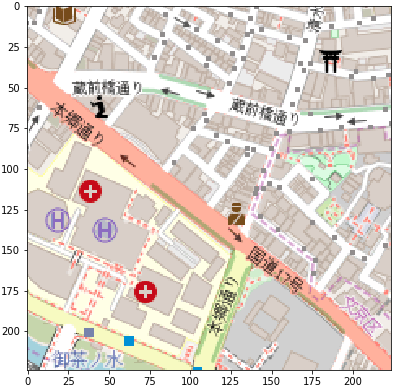}
        \caption{labeling 1 for classes of buildings more, highway, water, river, railway, rail station, park, playground, road more, airport, grassland, and labeling 0 for the others}
    \end{subfigure}\hfill
    \begin{subfigure}[t]{.48\linewidth}
        \centering
        \includegraphics[align=t, width=1.0\linewidth]{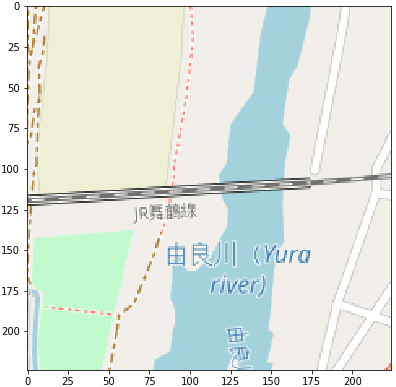}
        \caption{labeling 1 for classes of buildings less, highway, water, river, railway, park, road some, trail, farmland, and labeling 0 for the others}
    \end{subfigure}
    \caption{Examples of Map Tile Images and Their Labels}
    \label{fig:map_labels}
\end{figure}

Investigation of the metadata reveals that the data distributions significantly differ among the 14 classes. Moreover, when we compare if two map-tile images are alike from the geospatial and semantic perspectives, the specific count number for most classes might not be crucial. Instead, whether the class exists is more relevant. Recognizing this fact, we use binary number $1/0$ to label the of existence/non-existence of the 12 classes, i.e., all classes excluding building and road, in the image. For building and road, due to their relatively much wider range of metadata than the other 12 classes, we use 3 non-overlapping sub-classes to cover each of them, where the sub-classes are defined by comparing the metadata against pre-determined thresholds. Specifically, the class ``building" is redefined to be 3 classes ``building less", ``building some", ``building more" for the count number to fall into the intervals $[0,3)$, $[3,60]$, $(60,\infty)$, respectively; similarly, the class ``road" is redefined to be 3 classes ``road less", ``road some", ``road more", based on if the count number falls into the intervals $[0,15)$, $[15,30]$, $(30,\infty)$, respectively. The values of the thresholds are carefully chosen so that the resulting sub-classes are relatively balanced. To summarize, with the definition above, we create 18 binary labels for the resulting 18 classes for each image. Such a 18-dimensional binary vector for each image is referred to as "meta label" throughout this work. In Fig. \ref{fig:map_labels}, we provide two examples for visualization.

\subsection{POI Information}

POI is another category of information of interest in location-based applications. In this work, for each map-tile image, we collect all the POI names under the key of ``amenity" after parsing the output of the OpenStreetMap Overpass API to JSON files. After counting how many number of each POI name, e.g., restaurants, convenience stores, post offices, etc., appears in one image, we obtain the counting numbers of each POI name for each image. Note that the POI names for each map-tile image could significantly differ. Therefore, for the map-tile datasets, we collect all the POI names for all the images, and then create a POI count vector for each image, indicating if each POI name exists or not.

\subsection{Summary of the Notations}

For the reader to better understand this work, we summarize the notation use in the following. We use $l$ to represent a location tuple (\emph{lat}, \emph{lon}), and $M_l$ to represent the the map-tile image centered at $l$. Also, we denote by use $\mathcal{L}$ the location set and $\mathcal{M}_{\mathcal{L}}$ the map-tile image set for $\mathcal{L}$. In this work, since ESLE takes images as input, we use $l$ and $M_l$ interchangeably, and $\mathcal{L}$ and $\mathcal{M}_{\mathcal{L}}$ (or $\mathcal{M}$ by dropping the foot index) interchangeably when no ambiguity is caused. For a dataset $\mathcal{M}$, we denote by $N=|\mathcal{M}|$ its cardinality. For brevity, we denote by $\mathcal{C}$ the set of the 18 (sub-)classes of meta label data that we introduced. In addition, we use ${\bf y}_n$ to represent the meta label vector of the $n^{th}$ image, where the $c^{th}$ element of ${\bf y}_n$, denoted by $y_{n,c}$, is the binary label, where the index $n,c$ are automatically defined after the data is well collected. Once again, our goal is to create ESLE, which transforms $M_l$ to a numerical vector ${\bf e}_l\in \mathbb{R}^E$, where $E$ is the dimension of the embedding vector space. Note that the transformation is a deterministic mapping once the model is established. For brevity, we denote this mapping function as ${\bf e}_l=f(M_l)$. Concatenating all ${\bf e}_l$'s into a compact form produces an $N\times E$ embedding matrix ${\bf E}$, each row corresponding to one ${\bf e}_l$. 

Except for using the letters $l$ and $M_l$ for a location tuple and the corresponding map-tile image, we generally use lower case $a$ and upper case $A$ to denote scalar, and use the bold-fold ${\bf a}$, ${\bf A}$ to denote vector, matrix, respectively. In addition, $\mathcal{A}$ represents a set, and $\|{\bf a}\|$ denotes the Euclidean norm of ${\bf a}$.


\section{ESLE Embedding Models}\label{sec:lov2vec}

In this section, we walk the reader through the methodology to build the ESLE embedding models and the experiment setup, followed by the performance evaluation of the model in terms of the statistical metrics. 

\subsection{Methodology}\label{sec:multi-label-cnn}

We formulate creating location embedding as a supervised-learning multi-label classification problem, and the goal is to learn the semantic labels in the meta label dataset. We employ a ResNet-based multi-label CNN model, owing to its well-known capability to learn image features. Specifically, we delete the last fully-connected layer and SoftMax activation and then add another two layers. The first layer is a $512\times E$ fully-connected linear transformation layer followed by SeLu activation \cite{klambauer2017selfnormalizing}, where $E$ is the desired dimension of the location embedding vector space. Also, since we expect the resulting embedding vector to have not only the non-negative-valued but real-valued entries, i.e., to span a larger space volume as to better describe the feature, SeLu is employed to offset the non-zero centered effect caused by the more popular Relu function. The second is an $E\times C$ fully-connected linear transformation layer to match with the $C=18$ labels of the meta label data and then followed by Sigmoid activation, which is used to detect if each of the $C$ labels exists or not in each map-tile image. 

Recall that the meta label $y_{n,c} \in \{0,1\}$ implies if the $c^{th}$ label exists in the $n^{th}$ map-tile image, and we denote by $p_{n,c}\in [0,1]$ the corresponding prediction probability from the model defined above. We choose the multi-label soft margin loss \cite{multi_label_soft_margin_loss} as the loss function below:
\begin{equation}\label{eqn:multi_label_soft_margin_loss}
L = - \frac{1}{NC} \sum_{n=1}^N \sum_{c=1}^C \bigg[y_{n,c} \log(p_{n,c}) + (1\!-\!y_{n,c}) \log\left(1\!-\!p_{n,c}\right)\bigg].
\end{equation}

Finally, the $E$-dimensional embedding vector can be obtained from the input to the last layer. Feeding a map-tile image dataset $\mathcal{M}$ to the model, we can obtain their embedding matrix ${\bf E}=f(\mathcal{M})$.

\subsection{Experimental Setup}\label{experiment}

We investigate the performance of the model defined in Sec. \ref{sec:multi-label-cnn} on the datasets of our interest and via hyper-parameters tuning. Specifically, we test ResNet-18, ResNet-50, ResNet-101 and ResNet-152 based multi-label CNN classifier. During the training, we minimize the loss function defined in (\ref{eqn:multi_label_soft_margin_loss}). Moreover, we choose the Adam optimizer, and the other parameters include: batch size = 64, epochs = 20, learning rate = 0.001, and $\alpha=1.0$. Finally, all the experiments are performed on the AWS p2.xlarge instance.

To show the effectiveness of the ResNet-based CNN model, we also train a multi-label classifier using a simple 7-layer CNN structure as the CNN baseline model, which consists of 3 convolutional layers and 4 fully-connected layers. As the image size at the input is $3\times 224\times 224$, we choose the convolutional kernels to be with the dimension ($9\times 5 \times 5$), ($18\times 5 \times 5$), ($36\times 3 \times 3$) in each layer, each followed by SeLu activation and a $2\times 2$ max-pooling layer, and choose the 3 linear layers with the dimension of $22,500\times 1,024$, $1,024\times 128$, $128\times E$, each followed by Selu activation. At the end, similar to our proposed method, we add an $E\times C$ linear layer followed by Sigmoid activation. Aside from these, all the other parameters remain the same.

\subsection{Results Analysis}\label{subsec:mesh_eval}

Since the meta label data is in-balanced for each label, we evaluate the obtained location embedding vectors by using statistical metrics including precision, recall, F1-score \cite{precisionrecall} and Matthews correlation coefficient (MCC) \cite{mcc}. Next, we perform a number of experiments on the datasets $\mathcal{M}$ and/or $\mathcal{M}_1$ for model evaluation.

\begin{table}[t]
    \centering
    \begin{tabular}{c|cccc}
    \hline\hline
    \multirow{2}{*}{Models} & \multicolumn{4}{c}{Label Evalutaion}                                              \\
    & Precision & Recall & F1-score & \multicolumn{1}{c}{MCC} \\ 
    \hline
    Statistical Baseline& 0.1615 & 0.2601 & 0.2436	& 0.1309 \\
    CNN Baseline & 0.3838 & 0.4776 & 0.4880	& 0.0455 \\
    ResNet-18& 0.6129 & 0.6659 & 0.6900	& 0.6609 \\
    ResNet-50& 0.6344 & 0.6908 & 0.7045	& {\bf 0.6886} \\
    ResNet-101& 0.6353 & {\bf 0.6918} & {\bf 0.7072}	& 0.6787 \\
    ResNet-152& {\bf 0.6358} & 0.6795 & 0.7067	& 0.6829\\
    \hline\hline
    \end{tabular}
    \vspace*{1mm}
    \caption{Evaluation on 6 models for multi-label prediction}
    \label{tab:base_18_50_101_152_eval}\vspace{-0.1in}
\end{table}

Firstly, we consider the four models based on ResNet-18, ResNet-50, ResNet-101 and ResNet-152 proposed in \ref{sec:multi-label-cnn} and compare them against the CNN baseline model described in Sec. \ref{experiment}. We also show the statistical baseline, where prediction is made on the test data just based on the arithmetic mean of each label in the training dataset. Since the goal is to find out the relatively best model, we train the models on $70\%$ of subset $\mathcal{M}_1$ and test the models on the other $30\%$ of $\mathcal{M}_1$. As shown in Table \ref{tab:base_18_50_101_152_eval}, it can be seen that the ResNet-based CNN models {\em all} significantly outperform the two baselines in \emph{all} the four metrics. While the four ResNet-based CNN models outperform the CNN baseline due to their deeper structures, it can be seen that among themselves the deeper models such as ResNet-101 and ResNet-152 do not necessarily perform better, due to many possible reasons such as the amount of the data used for training and testing. Thus, considering the significantly greater amount of time and higher complexity for training deeper ResNet-based models, we decide to use ResNet-18 for further experiments. Another unimportant but interesting observation is the MCC comparison between the two baselines. While the CNN baseline significantly outperforms the other in precision, recall and F1-score, a bit surprisingly it is worse in MCC. An intuition explanation is that the MCC evaluates the prediction based on the extent of random guess, and it does not focus on characterizing how well a model learns the label 1. As they both tend to be 0, we do not need to pay much attention to the baselines in the rest of this paper.

\begin{table}[t]
    \centering
    \begin{tabular}{c|cccc}
    \hline\hline
    \multirow{2}{*}{Test Data} & \multicolumn{4}{c}{Label Evaluation}                                                    \\
        & Precision & Recall & F1-score  & \multicolumn{1}{c}{MCC} \\ 
    \hline
    Original Images               & 0.7415    & 0.6824 & 0.6975 & 0.6665 \\
    Rotated Images       & 0.7372    & 0.6764 & 0.6918 & 0.6601 \\ 
    \hline\hline
    \end{tabular}
    \vspace*{1mm}
    \caption{Evaluation on model robustness via image rotation}
    \label{tab:rotated_eval}
\end{table}

Next, we are interested in the model robustness. In particular, after we split the dataset $\mathcal{M}_1$ into the training set and the test set, we form a rotated test dataset by clock-wisely rotating each map-tile image by a random degree uniformly drawn from $\{0^o, 90^o, 180^o, 270^o\}$. After training the model using the training set, we test the model on both the test set and the rotated test set. If the model is not robust, then the test performance on the rotated test set should be worse. However, a bit surprised, as shown in Table \ref{tab:rotated_eval}, the result comparison reveals that there is no significant difference between the performance of the model on both test sets.  

\begin{table}[t]
    \centering
    \begin{tabular}{c|cccc}
    \hline\hline
    {Data Percentage} & \multicolumn{4}{c}{Label Evaluation}                                                    \\
    Used for Training & Precision & Recall & F1-score & MCC \\ 
    \hline
    7\%               & 0.8979 & 0.8474 & 0.8606 & 0.8488\\
    21\%              & 0.9021 & 0.8450 & 0.8625 & 0.8518\\ 
    35\%              & 0.9051 & 0.8570 & 0.8707 & 0.8602\\ 
    49\%              & 0.9071 & 0.8608 & 0.8739 & 0.8635\\ 
    63\%              & {\bf 0.9145} & {\bf 0.8618} &{\bf 0.8782} & {\bf 0.8684}\\ 
    90\%              & 0.9059 & 0.8586 & 0.8723 & 0.8619\\
    \hline\hline
    \end{tabular}
    \vspace*{1mm}
    \caption{Evaluation Results on Multi-Label ResNet-18 Model with Different Size of Training data}\vspace{-0.15in}
    \label{tab:ResNet18_eval}
\end{table}

Finally, we explore how much percentage of the data from $\mathcal{M}$ is needed for training the model to achieve certain satisfactory evaluation performance on the test dataset. Towards this task, we use $90\%$ and the other $10\%$ of $\mathcal{M}$ to form the training and test subsets, respectively. As shown in Table \ref{tab:ResNet18_eval}, we train the ResNet-18-based CNN model on $7\%$, $21\%$, $35\%$, $49\%$, $63\%$ and $90\%$ (i.e., all the training subset) of the entire dataset, and we test the model on test data subset. Note that from $7\%$ to $90\%$ in sequence, the dataset used for training the model is a randomly-drawn subset of the successive dataset, so as to remove the uncertainty caused by the random selection of the map-tile images during training. After training the model for 20 epochs for each experiment, we evaluate the model in four statistical metrics including precision, recall, F1-score, and MCC. It turns out generally the more amount of data, the better model performance, and the option of $63\%$ produces the relatively best performance for {\em all} evaluation metrics. The option of $90\%$ does not perform better than the $63\%$ option due to the restriction of the epoch number. In fact, based on our investigation, more epoch numbers can bring only little marginal improvement at the cost of much higher complexity. Thus, for the goal of concept proof and the training efficiency, we use the $63\%$ option to train the model and to create location embedding in the rest of this paper.

\section{Semantic Analysis}

To further interpret the semantic meaning of ESLE, we compare ESLE against the well-known Word2Vec for word embedding. In Word2Vec, we can perform addition and subtraction operations to explore the semantic relationship between the words, because word embedding has semantic information \cite{word2vec}. The additive property comes from the fact that the word embedding vectors represent the \emph{distribution of the context} in which a word appears, and they are \emph{linear} with model input and \emph{non-linear via softmax} with model output. Thus, linear addition in the word embedding vector space is related to the \emph{production of the probability} computed by the model, which can carry word semantics via probability in the probability space. In contrast in ESLE, the location embedding vector represents the \emph{distribution of each feature} existing in the map-tile image. The location embedding vector is \emph{non-linear} with model input and \emph{non-linear via sigmoid} with each label of model output. Linear addition in the location embedding vector space is related to the \emph{Boolean ``or" operation} of the model outputs. Thus, the next question is whether our location embedding generally carries semantic information. To answer this question, we first examine the addition and subtraction compositionality via visualizing some examples. Afterwards, we develop a systematic approach to extract the feature vectors of the 18 labels defined in Sec. \ref{subsec:meta}.

\subsection{Addition and Subtraction Compositionality}\label{subsec:semantic_add}

\begin{figure}[t]
    \centering
    \includegraphics[width=\linewidth,keepaspectratio]{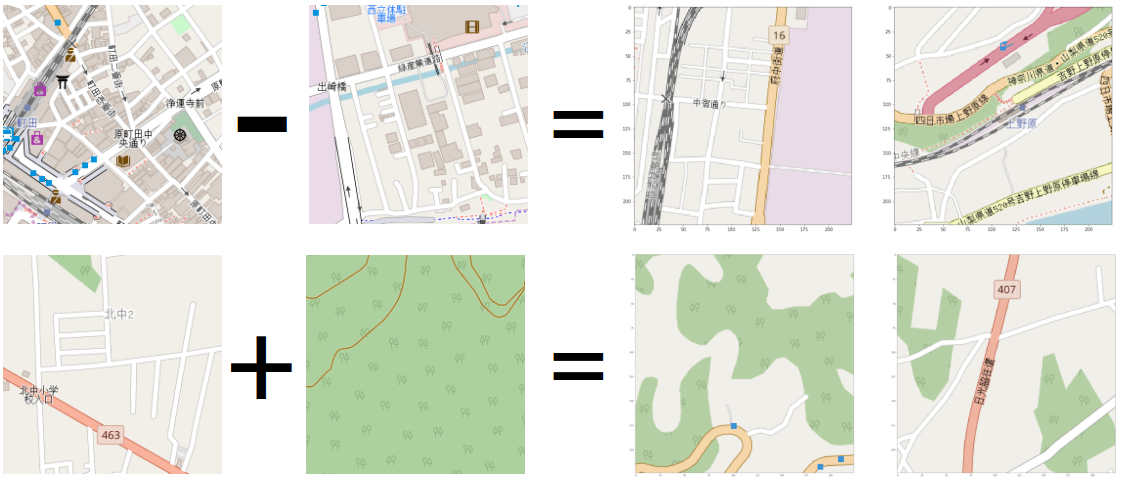}
    \vspace{-0.1in}
    \caption{Examples of the vector compositionality via element-wise addition and subtraction}
    \label{fig:add_sub}\vspace{-0.1in}
\end{figure}

Take Fig. \ref{fig:add_sub} as an example. We randomly pick 4 map-tile images (left to the equal sign) and perform element-wise addition and subtraction on their embedding vectors generated by ESLE. Then for the resulting two vectors, we search over all rows of the embedding matrix ${\bf E}$, retrieve the 2 nearest neighbors for each based on the cosine similarity, and show the corresponding map-tile images. In the first row, in the 2 images on the left-hand side of the first row in Fig. \ref{fig:add_sub}, both the minuend and the subtractor contain many buildings. After the subtraction operation, the resulting images do not contain or contain very few buildings; in the second row, in the two images on the left-hand side of the second row, one addend has freeway and no green land, and the other addend has much green land and no freeway. After the addition operation, the resulting two images contain both freeway and green land. Clearly, for both addition and subtraction operations, the retrieved images are quite meaningful and consistent with our interpretation. It can be seen that the meta information elements on the images on the right-hand side of the operation are added to (or deleted from) the images on the left-hand side.

\subsection{Label Feature Vector Visualization}

While Fig. \ref{fig:add_sub} can be used for the visualization verification purpose, it is necessary to consider a more general and universal approach for verification. Inspired by the face attributes transfer in \cite{gardner2015deep, xiao2018elegant, upchurch2017deep}, we develop a systematic approach to extract class feature vector. The main idea is briefly stated as follows:
\begin{enumerate}
    \item For each of the 18 classes, we form a number of groups in such a way that each contains two sets of map-tile images whose meta label vectors only differ in the corresponding class. For example, for class 1, in a group, the images in the first set all have meta label vector $[1,0,\cdots,0]$ and the images in the other set all have meta label vector $[0,0,\cdots,0]$. Since the weights of all the other 17 classes are all 0, we denote this group by Interfering Class Weights, ICW $= 0$; if ICW $\neq 0$, based on various possible combinations in $\mathcal{M}$, we readily form all the other groups for each possible ICW $=1,2,\cdots, 17$.
    
    \item We perform the subtraction operation between the embedding vectors of the images in the two sets of each group, aiming to remove all the other class information. Thus, the resulting vectors should carry the feature of the class of interest. Note that in each group, performing the subtraction operation between every two vectors (each from one set) would make the computation quite expensive. Instead, we only perform the operation for $\min(C_1, C_2)$ times where $C_1,C_2$ are the cardinality of the corresponding two sets, and each time the two embedding vectors are randomly selected from their sets, respectively, and without repetition (or replacement). 
    
    \item We analyze the distribution of the resulting feature vectors for all the 18 classes. To better visualize the distribution of the $E$-dimensional vectors, we perform PCA and Uniform Manifold Approximation and Projection (UMAP) \cite{mcinnes2018umap} to obtain their 2-dimensional representations. 
\end{enumerate}

\begin{figure}[t]
    \centering
    \includegraphics[align=t, width=1.0\linewidth]{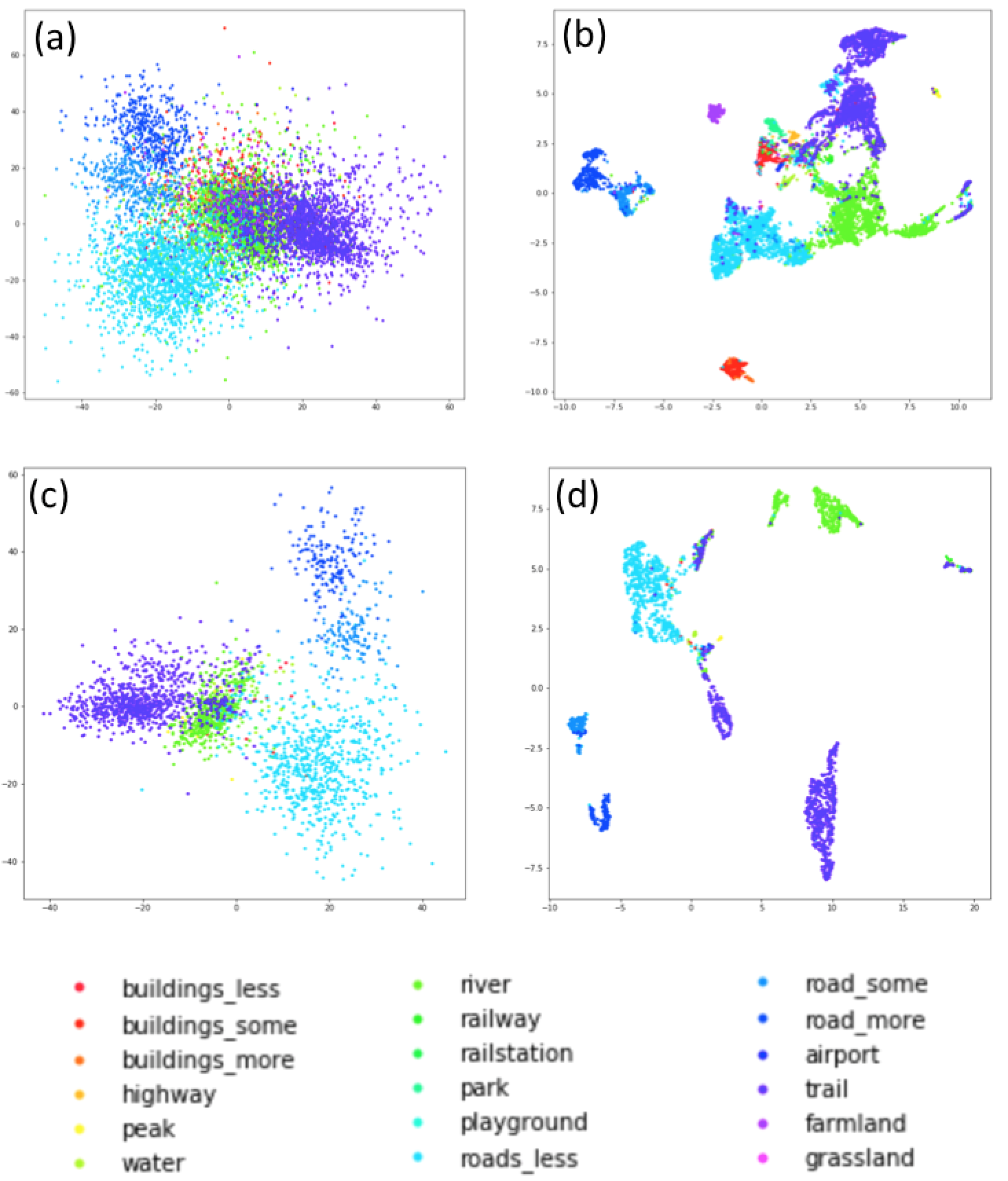}
    \caption{Visualization of the $C=18$ feature vectors: (a) PCA for (Interfering  Class  Weights) ICW $=1$, (b) UMAP for ICW $=1$, (c) PCA for ICW $=0$, (d) UMAP for ICW $=0$.}
    \label{fig:semantic2}\vspace{-0.1in}
\end{figure}

In Fig. \ref{fig:semantic2}, we visualize the feature vectors of the $C=18$ labels after performing PCA and UMAP for ICW$=1$ and ICW$=0$, respectively. Several interesting observations can be made. First, the clusters for ICW = 0 outperform those for ICW = 1 due to the class feature entanglement for ICW = 1. 
That is, those images with ICW = 1 carry much noise (of other class features) that cannot be completely removed after performing subtraction, thus making them harder to separate than ICW = 0. Second, the number of classes visible for ICW = 0 is less than that for ICW = 1, as the in-balanced class dataset excludes all the feature vectors for 4 classes for the group of ICW = 0. Finally, UMAP separates the clusters better than PCA, as UMAP considers the local similarity as well. To this end, the clustering effect can be evaluated by statistical metrics, such as Calinski and Harabasz (CH) index (we omit it here due to the absence of any baseline for comparison). Since the feature vectors are obtained from the embedding vectors via subtraction operation, we conclude that our location embedding carries semantic information.

\section{Application of ESLE to Bike Share Services}

After obtaining the location embedding, it is necessary to explore if it can help resolve any practical problem, like Word2Vec for natural language processing. In this section, we investigate if ESLE sheds light on the problem of seeking new ports for NTT DOCOMO's bike share services. There are 3 datasets that we use in this section:
\begin{enumerate}
    \item The first set $\mathcal{D}_1$ consists of the 1,622 locations of the service ports that NTT DOCOMO had established from 7/3/2015 to 7/22/2019, and the business operation starting date of each port \cite{share_bike_website}.
    \item The second set $\mathcal{D}_2$ consists of the 828 locations of the service ports operated in the Tokyo metropolitan area from Jan. 1, 2019 to Oct. 31, 2019, and also the hourly aggregated numbers of the bikes borrowed from and returned to each port. 
    \item The third set $\mathcal{M}_2$ is a subset of $\mathcal{M}_1$ defined in Sec. \ref{subsec:data_describe}. Specifically, among the 25,600 images of $\mathcal{M}_1$, we exclude any image of the locations within a certain distance to the existing ports. To expand the services, this constraint restricts our attention into the areas that are not too close to any of the existing service ports. As an example, we choose the distance threshold as 2.3 km, which is the minimum $10\%$ pair-wise distance between any two of the existing ports. This process removes 2,766 locations from $\mathcal{M}_1$ and the remaining 22,734 map-tile images form $\mathcal{M}_2$.
\end{enumerate}

Before illustrating how to use ESLE to seek new service ports, we first need to show evidence why it is possible. We begin with identifying the strong similarities of the existing service ports in the embedding vector space.

\subsection{Identification of the Existing Ports}\label{sec:sharebike_similarity}

Consider the existing 1,622 service ports in $\mathcal{D}_1$ in operation. Although they are broadly distributed over a very wide range including a number of cities, based on map-tile image observation they have a lot of attributes in common, such as many roads, many buildings, close to train stations. This is consistent with the common sense that bike share services should be highly desired in places with these attributes. Considering the distribution of their embedding vectors in the embedding vector space, we expect their envelops to constitute a subspace with relatively small volume, i.e., the embedding vectors of the existing 1,622 ports would be relatively close to one another. To verify this, we consider the distribution of all the embedding vectors of those in $\mathcal{D}_1$ and $\mathcal{M}_2$. In particular, let us investigate the similarities among the existing ports.  

\subsubsection{The Problem Formulation}

We consider the embedding vectors in dataset $\mathcal{D}_1$, where the ports were deployed over 49 months. For brevity, we denote by $\mathcal{E}_{D_1}(t), t=0,1,\cdots,48$ the set of embedding vectors of the corresponding ports deployed in the $t^{th}$ month. Also, we denote by $\mathcal{E}_{D_1}(t^-)=\cup _{i=0}^t\mathcal{E}_{D_1}(i)$ and $\mathcal{E}_{D_1}(t^+)=\cup _{i=t+1}^{48}\mathcal{E}_{D_1}(i)$ the embedding vectors of the ports deployed up to $t$ and after $t$, respectively. Based on $\mathcal{E}_{D_1}(t^-)$, we are interested if $\mathcal{E}_{D_1}(t^-)$ can be well characterized by the embedding vectors $\mathcal{E}_{M_2}$ of $\mathcal{M}_2$. While various approaches can be used for the question above, here we take a simple method to demonstrate the effectiveness of the embedding vectors. Specifically, we label the embedding vectors in $\mathcal{E}_{D_1}(t)$'s with 1 and those in $\mathcal{E}_{M_2}$ with 0. Then we train and test a binary classifier on $\mathcal{E}_{D_1}(t^-)$ and $\mathcal{E}_{M_2}$, and evaluate the performance of the classification results. 

\subsubsection{The Approaches and Evaluation Results}

\begin{figure}[t]
    \centering
    \includegraphics[align=t, width=1.0\linewidth]{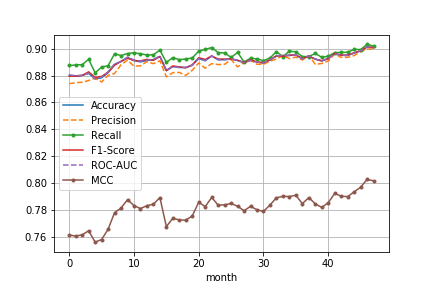}
    \caption{Classification performance evaluation based on the data up to each month since July 2015}
    \label{fig:similarity}\vspace{-0.1in}
\end{figure}

Since $|\mathcal{E}_{D_1}(t^-)|$ readily grows from $170$ to $1,622$ and the number is not large, we use Logistic Regression as the classifier. Moreover, the 1/0 labels for $\mathcal{E}_{D_1}(t^-)$ and $\mathcal{E}_{M_2}$ are highly skewed. To deal with it, for each $t$ ranging from 0 to 47, we uniformly draw $|\mathcal{E}_{D_1}(t^-)|$ (location) embedding vectors from $\mathcal{E}_{M_2}$ without repetition for a total of $T=200$ samples. Next, for each sample, we further randomly split the data of each class into training set and testing set based on $70\%/30\%$ splitting. Finally, for each $t$, we train $T$ models on the training data of the $T$ samples, and aggregate the statistical performance metrics on  the corresponding test dataset. In Fig. \ref{fig:similarity}, we show the results of the 6 metrics for $E=512$, including accuracy, precision, recall, F1-score, ROC-AUC and MCC. It can be seen that they \emph{all} are relatively stable. Also, they readily improve and converge with less and less variance, as more data comes into play over time. 

Besides this experiment, we also perform another experiment of using the $T$ models trained based on $\mathcal{E}_{D_1}(t^-)$ to score the probability, i.e., the output of the Sigmoid function, of the classifier with the vectors $\mathcal{E}_{D_1}(t^+)$ as the input. The experimental result shows that the score readily improves from 0.85 to 0.92 and the standard deviation reduces from 0.25 to 0.18 (based on the $T=200$ samples), as $t$ increases from 0 to 47. Thus, the existing ports are close to one another in the embedding vector space, 
and can be used for picturing the future new ports. 

With this new evidence, we investigate the problem of seeking new ports in the next two sections.

\subsection{How to Seek New Ports for Business Growth?}

Seeking new ports for bike share services is a complicated business decision-making problem, as analyzing the {\em dynamic} environment information (i.e., traffic flow, human density, security, and space) and operation performance (i.e., revenue, cost, profit, and their growth potential) is necessary. On the other hand, location embedding, geospatial and POI information, are all highly associated with the same map environment. Hence, the location embedding, carrying map-tile image information, could be useful to seek new ports. In this section, we show how to apply location embedding to this problem and how to evaluate the recommendation results. 

\subsubsection{The Problem Formulation}
Clearly, the task is not a traditional classification problem, as we do not have labels for multiple classes regarding the ports. However, we can still formulate it as a supervise-learning binary classification problem with the following method. Consider choosing a couple of ports from $\mathcal{M}_2$ as an example. Specifically, we first label all the 1,622 existing ports with 1, implying they are ``good", regardless its actual values in business. As shown in Sec. \ref{sec:sharebike_similarity}, they do have strong similarities among themselves. Then we label the 22,734 candidates in $\mathcal{M}_2$ with 0, meaning ``not good", even if some of them are ``good". From both business operation and map observations, since only a few candidates would finally be the target (e.g., selecting even no more than $1,000$ from the 22,734 ``not good" candidates would still leave the other more than $95\%$ ``not good"), we can use a binary classifier to approximately find the boundary between the 1/0 classes.  Thus, seeking new ports is equivalent to selecting a few ``not good" candidates that are very close to the ``good" ports, i.e., finding out the \emph{false negative} candidates most close to the ports with label 1, in the embedding vector space.

\subsubsection{The Approaches and Evaluation Results}

Similar as in Sec. \ref{sec:sharebike_similarity} and for $E=512$, we randomly draw 1,622 vectors from $\mathcal{E}_{D_2}$ without repetition, so as to make the data with 1/0 labels balanced. To reduce the uncertainty caused by random sampling, we independently sample $\mathcal{E}_{D_2}$ for $T=100$ times. Also, considering the computation complexity, small data-scale, and simplicity, we again use logistic regression as the classifier. Based on our experiments with $70\%/30\%$ for train/test splitting, the resulting accuracy is $0.9055$ with the standard deviation $0.0072$ over $T=100$ samples, suggesting logistic regression is a very stable model\footnote{We also experiment with other models including SVM, Spectral/K-Means Clustering, Decision Tree, MLPNN, and even the best of their performance results is not significantly better.}.

Second, we consider $E$ takes other values of 8, 64, 128, 256, 512, respectively. Intuitively, the larger $E$, the more information carried by the embedding vector. However, a larger $E$ also means more expensive computation and higher risk of over-fitting. Thus, we evaluate the performance metrics for all the choices for $E$. To show the effect of the embedding vectors, we also provide the results based on using ``Metadata", ``Meta label", ``POI" and the combined of these three. Table \ref{tab:POI_Label_Meta_bike_eval} demonstrates that $E=512$ performs the best. 

Next, we compare ESLE against Tile2Vec \cite{tile2vec}. Both models are trained by using 63\% of $\mathcal{M}$ described in Sec. \ref{subsec:mesh_eval} and the same parameters introduced in Sec. \ref{experiment}. As shown in Table \ref{tab:Triplet_Multilabel_bike_eval}, ESLE performs the best in all three statistical metrics, and ESLE saves the amount of the training time, in minutes per epoch, by $71.43\%$. In addition, we also consider a unsupervised-learning model -- AntoEncoder \cite{Goodfellow-et-al-2016} -- for performance comparison. Here, we directly employ the structure of the AutoEncoder developed in \cite{segnet} but make a few changes in the parameter values. Specifically, in the decoder, the number of (input, output) channels is $(3, 64)$ for module 1, $(64, 64)$ for module 2, 3, 4, $(64, 512)$ for module 5. After that we insert a convolutional layer with 512 $(7\times 7)$ filters to retrieve the embedding vector. The decoder is like a mirror of the encoder. As shown in Table \ref{tab:Triplet_Multilabel_bike_eval}, AutoEncoder performs much worse than both ESLE and Tile2Vec in all metrics.

\begin{table}
    \centering
    \begin{tabular}{c|ccc}
    \hline\hline
        Input Data & Accuracy & F1 score & AUC-ROC \\
    \hline
        POI & 0.8678 & 0.8652 & 0.8713 \\
        Meta Label & 0.8967 & 0.8967 & 0.8967 \\
        Metadata & 0.8839 & 0.8790 & 0.8840 \\
        Combined$^\dag$ & 0.8967 & 0.8981 & 0.8985 \\
        8-d Emb & 0.8790 & 0.8860 & 0.8782 \\
        64-d  Emb & 0.8928 & 0.8958 & 0.8938 \\
        128-d Emb & 0.8894 & 0.8927 & 0.8904 \\
        256-d Emb  & 0.8872 & 0.8906 & 0.8882 \\
        512-d Emb & {\bf 0.9055} & {\bf 0.9055} & {\bf 0.9055} \\
    \hline\hline
    \end{tabular}
    \vspace{0.05in}
    \caption{Bike share port prediction (Combined$^\dag$ denotes the concatenated use of POI, Meta Label and Metadata)}
    \label{tab:POI_Label_Meta_bike_eval}
\end{table}

\begin{table}[t]
    \centering
    \begin{tabular}{c|cccc}
    \hline\hline
        Models & \makecell{Training \\ Time}  & Accuracy & F1 score & AUC-ROC \\
    \hline
        AutoEncoder & 12h 30m & 0.7548 & 0.7625 &  0.7589 \\
        Tile2Vec\cite{tile2vec} & 1h 42m & 0.8868 & 0.8902 & 0.8876 \\
        ESLE & {\bf 32m} & {\bf 0.9055} & {\bf 0.9055} & {\bf 0.9055} \\
    \hline\hline
    \end{tabular}
    \vspace{0.05in}
    \caption{Bike share prediction comparison among different embedding models}
    \label{tab:Triplet_Multilabel_bike_eval}\vspace{-0.1in}
\end{table}

\subsubsection{New Port Recommendation}

To this end, given the embedding size $E$, we can find out the top-$M^*$ locations/map-tile images for recommendation, denoted by $\mathcal{R}_{M^*}$, which is a function of $P=$ ($E$, location embedding, candidate ports, classification model, $M^*$). However, if we pick another $P$, we would have another $\mathcal{R}_{M^*}$. While one can create many $P$'s based on different settings, it is not clear which $P$ is more efficient than the others. On the other hand, one cannot expect every location in $\mathcal{R}_{M^*}$ to be promising. To increase the reliability of the recommendation, we choose $K$ different $P$'s, say $P_{(k)}$ to obtain their corresponding $\mathcal{R}_{{M^*}(k)}=f_r(P_{(k)})$ for $k=1,2,\cdots,K$. Then we take their intersection $\mathcal{R}_{\bar{P}}=\cap_{k=1}^K \mathcal{R}_{M^*(k)}$ as the final answer. 

While each $\mathcal{R}_{{M^*}(k)}$ is with size ${M^*}$, it is of interest to see the cardinality of $\mathcal{R}_{\bar{P}}$. Consider two extreme cases: 
\begin{itemize}
\item $|\mathcal{R}_{\bar{P}}|$ is close to ${M^*}$: in this case, all $\mathcal{R}_{{M^*}(k)}$'s are alike, and thus the embedding mapping function $f(\cdot)$ tends to be many-to-one mapping and relatively stable regardless of the input. Thus, it suffices to use the smallest $E$. However, as shown in Table \ref{tab:POI_Label_Meta_bike_eval}, location embedding vectors with larger $E$ is more informative. Thus, we do not expect to see this extreme. 
\item $|\mathcal{R}_{\bar{P}}|$ tends to be $0$: in this case, every two $\mathcal{R}_{{M^*}(k)}$'s differ a lot, implying the intersection $\mathcal{R}_{\bar{P}}$ does not increase the reliability of the recommendation. Thus, we do not prefer to see this extreme as well. 
\end{itemize}
Generally, we expect the carnality of a desired recommendation list $\mathcal{R}_{\bar{P}}$ not to be close to both $0$ and $M$. 

\begin{figure}[!t]
\centering
\includegraphics[width=3.3in]{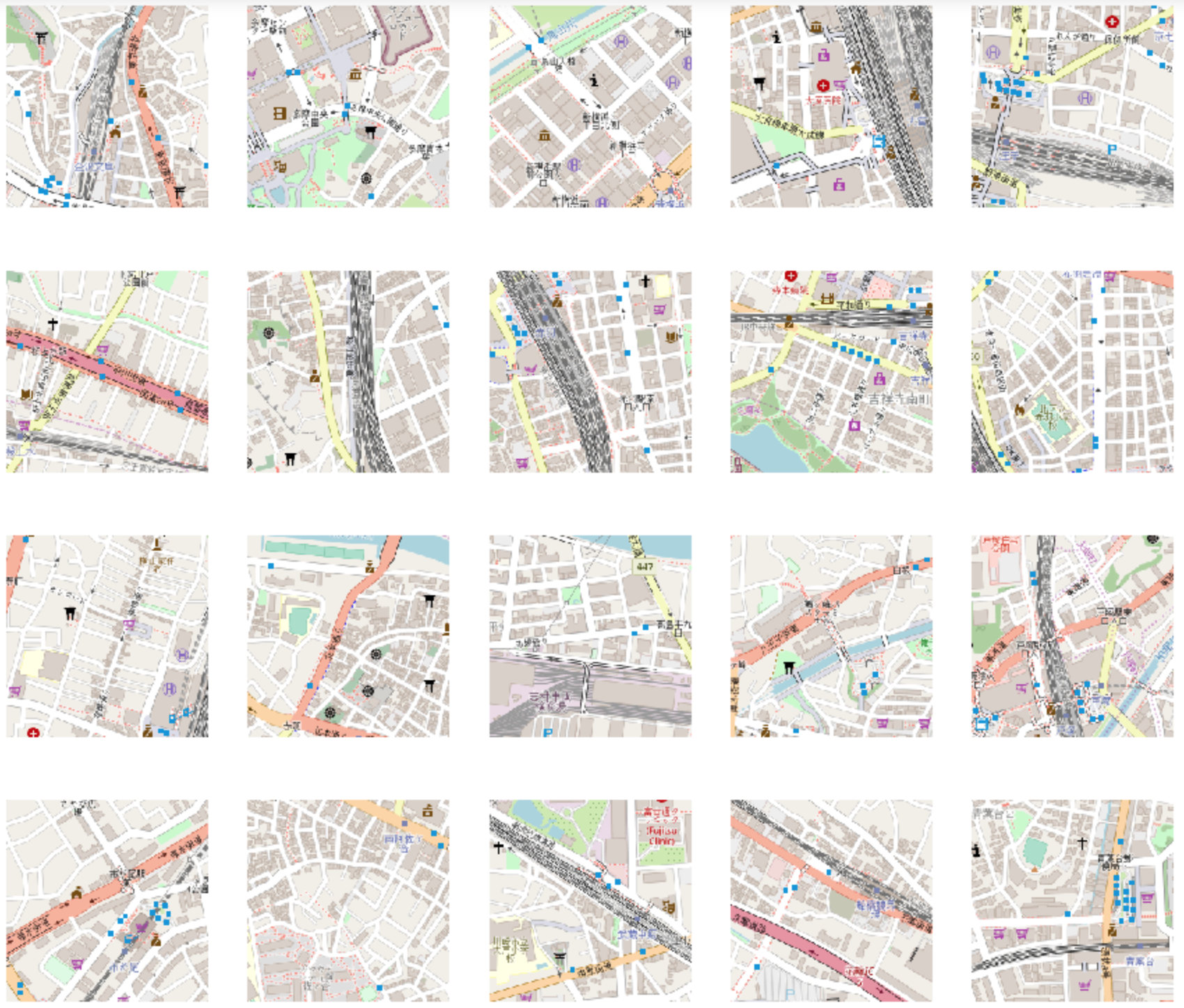}
\caption{The 20 recommended sites for ${M^*}=100$}
\label{fig:recommendation}
\end{figure}

In our experiments, we choose $K=4$, and the corresponding $P_{(k)}$'s are given by: $P_{(1)}$ (embedding vectors with $E=512$ only), $P_{(2)}$ (embedding vectors with $E=512$, metadata, meta label), $P_{(3)}$ (embedding vectors with $E=128$ only), and $P_{(4)}$ (embedding vectors with $E=128$, metadata, meta label). By choosing $M$ from 100 to 2,000, we show the $|\mathcal{R}_{\bar{P}}|$ in Table \ref{tabel_recommendation}. It can be seen that, first, the relative cardinality of $|\mathcal{R}_{\bar{P}}|$ compared to ${M^*}$ increases when ${M^*}$ grows, which implies that the four sets $\mathcal{R}_{{M^*}(k)}$'s tend to agree more among themselves; second, the ratio $|\mathcal{R}_{\bar{P}}|/{M^*}$ ranges over $(0.2, 0.53)$, consistent with our expectation. For example, if we pick ${M^*}=100$, then the top-20 location/map-tile images are recommended as shown in Fig. \ref{fig:recommendation}. It can be seen that each image has the features including railways, a lot of roads, a lot of building, etc., which are actually very similar to the map-tile images centered at the existing ports.

\begin{table}[t]
\centering
\begin{tabular}{c|cccccccc}
\hline\hline
$M^*$ & 100\! & 200\! & 400\! & 600\! & 800\! & \!1,200\! & \!1,600\! & \!2,000\!\\
\hline
$|\mathcal{R}_{\bar{P}}|$ & 20\! & 58\! & 140\! & 250\! & 350\! & 575\! & 801\! & \!1,060\!\\
\!\!$|\mathcal{R}_{\bar{P}}|/M^*$ \!\!\!\!\!\!& 0.2\! & 0.29\! & 0.35\! & 0.417\! & 0.438\! & 0.479\! & 0.5\! & \!0.53\! \\ 
\hline\hline
\end{tabular}
\vspace{0.05in}
\caption{The number of map-tile images agreed by all the four models (the intersection of their top-${M^*}$ list sets)}
\label{tabel_recommendation}
\end{table}

\subsubsection{Evaluation of the Recommendation}

\begin{table}[t!]
\centering
\begin{tabular}{c|cc}
\hline\hline
POI names & 1,622 existing images & 20 new images \\
\hline
restaurant & 11.9018 & 16.15 \\
cafe & 4.3412 & 4.85 \\
pub & 3.8479 & 5.85 \\
fast food & 2.9636 & 5.60 \\
vending machine & 2.8848 & 5.05 \\
bench & 1.9739 & 2.00 \\
bar & 1.8273 & 1.30 \\
post box & 1.7285 & 2.00 \\
telephone & 1.4309 & 2.55 \\
parking & 1.2352 & 2.35 \\
toilets & 1.1133 & 1.20 \\
bank & 1.0564 & 3.15 \\
pharmacy & 1.0473 & 3.25 \\
place of worship & 1.0170 & 0.95 \\
dentist & 0.8679 & 2.35 \\
post office & 0.7430 & 0.75 \\
bicycle parking & 0.6752 & 1.15 \\
doctors & 0.6230 & 2.45 \\
drinking water & 0.6036 & 0.45 \\
parking entrance & 0.5739 & 0.45 \\
kindergarten & 0.5697 & 0.60 \\
public building & 0.5424 & 0.25 \\
social facility & 0.5309 & 0.65 \\
police & 0.4709 & 0.95 \\
atm & 0.4448 & 1.45 \\
\hline\hline
\end{tabular}
\vspace*{1mm}
\caption{The most significantly important 25 POI names of the existing 1,622 ports and the POI count means over the existing 1,622 ports and the 20 recommended new ports}\vspace{-0.1in}
\label{tab:poi}
\end{table}

Besides visualization via Fig. \ref{fig:recommendation}, can we analytically evaluate $\mathcal{R}_{\bar{P}}$? Since the ground truth is not available before operating the business for a period, it is very difficult if not impossible. Of course, with more commercial business operation data, one could make the predict with higher confidence, but what evaluation metrics to use is still not straightforward. To solve these difficulties, we propose to leverage the POI information, which is not directly used when we train ESLE and perform binary classification. In addition, it is reasonable for people to use bike share services around places such as restaurant, park, bus station, etc., which essentially belong to POI information. Thus, we compare the POI information of the 20 recommended ports in Fig. \ref{fig:recommendation} against those existing 1,622 ports.  

From the 20 recommended map-tile images $\mathcal{R}_{\bar{P}}$ and the map-tile images of the 1,622 existing ports in $\mathcal{D}_1$, there are a total of 204 POI names. After calculating their mean counts for the 20 recommendation ports and $\mathcal{D}_1$ individually, we form two count mean vectors $V_{1622}$ and $V_{20}$ for $\mathcal{D}_1$ and $\mathcal{R}_{\bar{P}}$, respectively. A little surprisingly, $V_{1622}$ and $V_{20}$ are strongly dependent. This can be verified via their correlation, entropy, and KL divergence (i.e., the relative entropy). Specifically, first, their correlation coefficient (the cosine similarity score) is $0.9753$, which implies that the location embedding is able to learn the underlying POI information. Second, their individual entropy is given by $H(V_{1622}) = 4.5544$, $H(V_{20}) = 4.6295$ bits, respectively, very close to one another, both much less than $5.0844$ bits, the entropy of the POI count mean distribution of all the map-tile images in $\mathcal{M}_2$. For their KL divergence, $D(V_{1622}\| V_{20})=0.3724$ bits and $D(V_{20}\| V_{1622})=0.3706$ bits only, respectively. In Table \ref{tab:poi}, due to the space limitation we only show 25 highest counts for $\mathcal{D}_1$, their POI names, and the corresponding count values for $\mathcal{R}_{\bar{P}}$. It highlights the importance of those 25 POI names, where the first 5 POI names with counts higher than 2.8, and especially restaurant has count as high as $11.9018$, which indicates those POI names tend to be the key attributes based on our model and could be used for seeking new ports.

\subsection{How to Seek Ports with Relatively High User Flow?}

In the prior two tasks, we use the location of the existing service ports only. In this section, we take a step forward to investigate whether the operation data of the existing service ports and their location embedding vectors have dependencies. Obviously, the existing 1,622 bike share service ports do not bring the same business value, as they were deployed to meet varying demands at different places. For example, the service ports deployed in the urban downtown areas and the suburban area would see rather different bike usages and bike borrowed/returned patterns. Intuitively, the ports with higher usages, i.e., the bike-user flow, would probably have greater business values. Note that the location embedding vectors carry semantic information and some geospatial endowment information, which is somewhat associated with the service usage or demand information. Thus, an interesting question naturally arises: Whether location embedding vectors of the service ports imply the bike-user flow with certain extent? 

\subsubsection{The Problem Formulation}

We use the dataset $\mathcal{D}_2$, introduced at the beginning of this section. Borrowing the idea of data pre-processing used in \cite{bikesharedemand}, we define a service port to be ``high flow" if more than 24 bikes are borrowed and returned in average per hour, and to be ``low flow" for otherwise. This definition splits the 828 ports as 473 ``higher flow" ports and 365 ``low flow" ports, labeled with 1 and 0, respectively. Then the question becomes how much we can tell the labels of the service ports based on their location embedding vectors only. Clearly, this is a binary classification problem. Again, although there are many other approaches of machine-learning problem formulation, we formulate this binary classification problem for simplicity, as our main purpose is to shed light on the application of location embedding.

\subsubsection{The Approaches and Evaluation Results}

\begin{table}[t]
    \centering
    \begin{tabular}{c|ccccc}
    \hline\hline
        Input Data & Accuracy \!\!&\!\! F1 score \!\!&\!\! AUC-ROC \!\!&\!\! MCC \!\!\\
    \hline
        Statistical Baseline & 0.5712 & 0.5712 & 0.5712 & 0 \\
        512-d Emb & 0.6224 & 0.5254 & 0.6065 & 0.1873\\
        Combined$^*$ & {\bf 0.6548} & {\bf 0.5923} & {\bf 0.6471} & {\bf 0.2723} \\
    \hline\hline
    \end{tabular}
    \vspace{0.05in}
    \caption{Bike share service port user flow prediction (Combined$^*$ denotes the concatenated use of 512-d Embedding, POI, and Metadata)}\vspace{-0.2in}
    \label{tab:flow_eval}
\end{table}

To keep the model as simple as possible, we again use logistic regression as the classifier. For each port, besides using its location embedding vector only, we can also consider the combined use with its POI vector and meta label vector for performance comparison. Similar as in the prior two sections, we split the data into the training set and the test set based on $70\%/30\%$ splitting. Then we train the classifier on the training set and evaluate the classifier on the test set. As shown in Table \ref{tab:flow_eval}, the performance evaluation results are performed based on using using "$E=512$-dimensional embedding vectors" only and all of location embedding vector, ``POI", ``Meta Label" as the input to the classifier. Compared to the the statistical baseline where the predication is made with with distribution-based random guess, the location embedding demonstrates its effectiveness. Although the statistical baseline is better in F1-score, it does not learn any underlying structure behind the data, as its MCC $=0$. In addition, even the metric values corresponding to the location embedding as the input are way below 1, we never expect them to be close to 1, because this is only an attempt to study whether the dependencies between location embedding vectors and the port usage exist. 

Once the dependencies exist, as shown in Table \ref{tab:flow_eval}, it would be worthy of exploring other smarter methods to better formulate the problem, pre-process the data, and characterize the underlying structure of the data. For example, when the ``POI" and ``Meta Label" vectors come into play as well, the three performance metrics in Table \ref{tab:flow_eval} significantly outperform that using location embedding only.

\section{Conclusion}

We develop a universal location embedding model -- ESLE -- to encode an map-tile image of a given location into a numerical vector with much lower dimensions. ESLE is built by using a ResNet-based multi-label classification CNN model with the metadata as the output label, and trained in the supervised-learning manner. The experiments indicate ESLE is relatively effective in general and outperforms several baseline approaches. Also, based on a systematic approach, we demonstrate that the resulting location embedding vectors carry both spatial geometry and semantic information. Finally, We apply the trained ESLE to a business problem for NTT DOCOMO's bike share services. Particularly, we attempt to show its values in understanding the underlying features of the existing service ports and thus in recommending new service ports for expanding the services in new areas. The initial results shown in this paper indicate our ESLE model could be a very promising and efficient tool for a wider range of location-based applications. The future work would include leveraging more operation data of bike share services to further examine ESLE and to make ESLE more powerful. Moreover, exploring other clever methods to create more efficient and interpretable ESLE is another important and interesting direction.

\bibliographystyle{ieeetr}
\bibliography{ref}


\end{document}